\documentclass[lettersize,journal]{IEEEtran}
\usepackage{amsmath,amsfonts}
\usepackage{algorithm}
\usepackage{algorithmicx}
\usepackage{array}
\usepackage[caption=false,font=normalsize,labelfont=sf,textfont=sf]{subfig}
\usepackage{textcomp}
\usepackage{stfloats}
\usepackage{url}
\usepackage{verbatim}

\usepackage{cite}
\usepackage{algpseudocode}
\usepackage{float}

\usepackage{graphicx}
\usepackage{subfig}

\usepackage[margin=1in]{geometry}
\usepackage{pgfplots}
\usepackage{tikz}
\usetikzlibrary{patterns}

\pgfplotsset{compat=1.18}
\usepackage{pgfplotstable}
\usepackage{xcolor}
\usepackage[margin=1in]{geometry}
\pgfplotsset{compat=1.18}
\usepackage[margin=1in]{geometry}
\begin{document}

\title{Causal Neighbourhood Learning for Invariant Graph Representations}

\author{Simi~Job,
        Xiaohui~Tao,
        Taotao~Cai,
        Haoran Xie, 
        and~Jianming~Yong 
\thanks{S. Job, X. Tao and T. Cai are with the School of Science, Engineering and Digital Technologies, University of Southern Queensland, Australia.}

\thanks{H. Xie is with the School of Data Science, Lingnan University, Hong Kong.}

\thanks{J. Yong is with the School of Business, Law, Humanities and Pathways, University of Southern Queensland, Australia.}
}


\maketitle

\begin{abstract}
Graph data often contain noisy and spurious correlations that mask the true causal relationships, which are essential for enabling graph models to make predictions based on the underlying causal structure of the data. Dependence on spurious connections makes it challenging for traditional Graph Neural Networks (GNNs) to generalize effectively across different graphs. Furthermore, traditional aggregation methods tend to amplify these spurious patterns, limiting model robustness under distribution shifts. To address these issues, we propose Causal Neighbourhood Learning with Graph Neural Networks (CNL-GNN), a novel framework that performs causal interventions on graph structure. \textit{CNL-GNN} effectively identifies and preserves causally relevant connections and reduces spurious influences through the generation of counterfactual neighbourhoods and adaptive edge perturbation guided by learnable importance masking and an attention-based mechanism. In addition, by combining structural-level interventions with the disentanglement of causal features from confounding factors, the model learns invariant node representations that are robust and generalize well across different graph structures. Our approach improves causal graph learning beyond traditional feature-based methods, resulting in a robust classification model. Extensive experiments on four publicly available datasets, including multiple domain variants of one dataset, demonstrate that \textit{CNL-GNN} outperforms state-of-the-art GNN models.
\end{abstract}


\section{Introduction}

Graph data contain complex and noisy connections, where a node's neighbours may influence its label causally or spuriously. Traditional Graph Neural Networks (GNNs) aggregate information from all neighbours uniformly or through attention mechanisms, which can indiscriminately include spurious correlations and overlook important causal patterns. Node features may also contain both causally informative and confounding attributes, making it difficult to disentangle them without domain-specific knowledge. These factors cause GNNs to overfit to spurious patterns, resulting in poor generalization under shifts in node features or graph structures \cite{zhu2021shift}. Real-world graphs are often noisy or incomplete. For instance, in citation networks, co-authorship edges may reflect social relationships rather than topical similarity, resulting in non-causal correlations and poor generalization. To address these challenges, a causally-grounded graph learning approach based on causal principles and invariant prediction is required to improve out-of-distribution generalization \cite{pearl2010introduction, sui2024invariant, wang2025causal, li2025causal}. 

Invariant Graph Learning (IGL) \cite{sui2024invariant} is a method developed to address invariance in graphs and it learns global invariant representations by simulating environments and modeling invariant confounders. Causality Inspired Invariant Graph LeArning (CIGA) \cite{chen2022learning} identifies class-level invariant subgraphs that capture causal label information for graph classification tasks under distribution shifts. Cluster Information Transfer (CIT) \cite{xia2023learning} uses cluster-based representation learning to achieve invariance under structure variations. Other approaches include causal graph contrastive learning with spectral augmentation \cite{mo2024graph} and Generative-Contrastive Collaborative Self-Supervised Learning (GCCS) \cite{zhang2025self} to learn invariant causal representations in heterogeneous information networks. 

 To address the limitations of existing approaches, we propose Causal Neighbourhood Learning with Graph Neural Networks (CNL-GNN), a framework that extends beyond uniform aggregation by performing causally-informed interventions at the neighbourhood and edge levels. By generating counterfactual neighbourhoods and adaptively perturbing edges through learnable masking, CNL-GNN dynamically identifies and preserves causally relevant graph structures and reduces the influence of spurious connections. This structural-level causal intervention enables the model to learn invariant node representations that generalize across varying graph structures, extending the scope of causal learning beyond feature-based methods. In contrast to prior work focusing on invariant node or subgraph representations through multi-environment simulation \cite{sui2024invariant, chen2022learning}, or cluster-based transfer \cite{xia2023learning}, our approach directly intervenes on the graph structure through Counterfactual Neighbourhood Generation, simulating alternative causal scenarios at the connectivity level. We further incorporate an adaptive, learnable edge importance mechanism to selectively perturb edges and disentangle causal from spurious relationships. This structural intervention combined with feature disentanglement enables more robust and causally consistent node classification under confounding and distribution shifts. \textit{Causal Neighbourhood Leaning} (CNL) facilitates invariant node representations through counterfactual neighbourhood reasoning \cite{wang2025improving, lee2025policy}. Traditional GNNs aggregate features from observed neighbours, whereas Causal Neighbourhood Learning simulates counterfactual scenarios in which a node is connected to a different set of neighbours. This enables the model to identify stable, causally relevant signals across contexts. By integrating counterfactual reasoning, interventional perturbations and invariant representation learning into neighbourhood aggregation, CNL provides a foundation for causally robust GNNs that generalize across domains. This research makes the following key contributions:
\begin{itemize}
    \item \textit{Counterfactual Neighbourhood Generation.}
We propose an approach that generates counterfactual neighbourhoods by explicitly intervening on node connectivity. This enables the model to simulate various causal scenarios at the structural level, enhancing its ability to disentangle true causal relationships from spurious correlations.

    \item  \textit{Adaptive Edge Perturbation with Learnable Importance Masking.}
    We introduce an adaptive edge perturbation technique that employs learnable importance scores to selectively mask and modify edges. This encourages the model to focus on causally relevant connections, improving robustness against noisy or spurious graph structures. This builds on the counterfactual neighbourhoods by using learned causal importance to guide perturbations.

    \item \textit{Attention-Based Edge Scoring for Causal Relevance Estimation.}
A tailored attention mechanism estimates edge importance scores to differentiate causal from non-causal connections, enabling targeted structural interventions that selectively perturb or preserve edges. This approach extends conventional attention models by facilitating invariant graph representations that effectively capture stable causal relationships. These scores form the basis of the adaptive masking mechanism and support the generation of meaningful counterfactuals.

\end{itemize}

These form an integrated framework that enhances causal understanding in graph domains by generating informative perturbations and focusing on causally meaningful structures.

\section{Related Work}

This section provides a brief overview of related studies in causal learning and graph neural networks.
\begin{figure*}[h]
\centering
\includegraphics[width=0.85\textwidth]{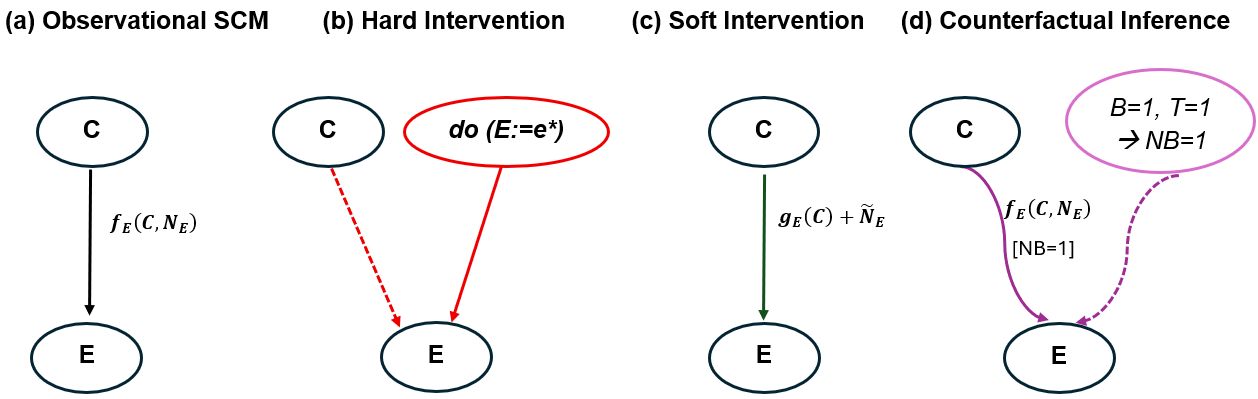}
\caption{Illustration of causal concepts using SCMs.
(a) cause \textit{C} influences effect \textit{E}.
(b) sets \textit{E} to a fixed value.
(c) modifies function or noise affecting \textit{E}. ($N{_E}$: noise)
(d) updates noise to evaluate alternative outcomes without changing causal structure. (T=1, B=1 : observed treatment and outcome
NB=1: inferred latent condition)}
\label{fig:process}
\end{figure*}

\subsection{Causal Learning}

Causal Learning involves modeling cause and effect relationships in data, enabling models to make robust and generalizable predictions by distinguishing true causal factors from spurious correlations. It has been applied in various domains including healthcare \cite{sanchez2022causal, ferrari2022causal}, economics \cite{celli2022causal}, social science \cite{manning2024automated} and finance \cite{chen2024probability, paladugu2025causal}. In healthcare, Ferrari \emph{et al.}~\cite{ferrari2022causal} used causal learning to analyze pandemic data. Sanchez \emph{et al.}~\cite{sanchez2022causal} applied it to clinical decision support systems by predicting outcomes based on specific treatments and assessing the effects of interventions. In the financial domain, Chen \emph{et al.}~\cite{chen2024probability} applied probabilistic causal inference to model the directional influence between loan amounts and interest rates, providing insights into market behaviour and policy impacts in credit market. In the same domain, Paladugu~\cite{paladugu2025causal} integrated causality into data pipelines enabling models to reason about interventions and counterfactuals, and enhancing robustness and decision quality in dynamic markets. In the following subsections, we briefly discuss causal learning under two main categories: (i) graph-based causal learning and (ii) counterfactual and interventional approaches.

\subsubsection{Graph-based Causal Learning}

Graph-based causal learning uses graphical models such as directed acyclic graphs (DAGs) to represent and identify causal relationships among variables. It uncovers the underlying causal structure from data and enables analysis of how changes or interventions affect outcomes. There are several graph causal learning approaches, including constraint-based methods such as \textit{PC} and \textit{FCI}, score-based methods such as \textit{GES} and continuous optimization methods such as \textit{NOTEARS}. Additionally, methods based on additive noise assumptions such as \textit{LiNGAM} are commonly used to uncover causal structures from data \cite{job2025exploring}.

Kalisch \emph{et al.}~\cite{kalisch2007estimating} analyzed the PC-algorithm for estimating the skeleton and equivalence class of very high-dimensional, sparse Gaussian DAGs, proving its uniform consistency as the number of nodes grows rapidly with sample size, and demonstrating its effectiveness on simulated data. Zhang \emph{et al.}~\cite{zhang2012abnormal} applied ICA-LiNGAM combined with SVAR to time-series driving behaviour data to uncover instantaneous and lagged causal relationships, advancing driving action analysis for assistance systems.

\subsubsection{Counterfactual and Interventional Approaches}

For reliable causal learning, two primary approaches can be used. Direct intervention involves systematically manipulating one or more variables to observe the resulting effects on other variables, thereby helping to identify and confirm causal relationships. Counterfactual reasoning infers causality by considering alternative scenarios to observed data, which may or may not involve actual interventions.

Rohbeck \emph{et al.}~\cite{rohbeck2024bicycle} introduced \textit{Bicycle}, a novel method for inferring cyclic causal graphs from i.i.d. and interventional data. By modeling causal effects in a latent space, it demonstrates superior performance on both simulated and real single-cell datasets. Shanmugam \emph{et al.}~\cite{shanmugam2015learning} designed a model for causal learning with interventions under Pearl’s Structural Equation Model with independent errors (SEM-IE), aiming to recover all causal directions in a graph using the minimal number of interventions, and introduced adaptive algorithms along with information-theoretic lower bounds. Causal interventions were also employed in developing a method that uses generative models to learn robust visual representations, improving generalization to out-of-distribution data \cite{mao2021generative}.

A causal disentangling framework was proposed by Li \emph{et al.}~\cite{li2024causal}, which learns unbiased causal effects by addressing inductive and dataset biases through counterfactual interventions. Zhou \emph{et al.}~\cite{zhou2024decision} proposed Decision Focused Causal Learning (DFCL), a framework that uses counterfactual reasoning to address key challenges in marketing budget optimization, and has been successfully deployed on a major food delivery platform. Thus, counterfactuals and interventions can be used either together or separately to enhance causal learning.

\subsection{Graph Neural Networks}

In recent times, Graph Neural Networks have been increasingly applied for causal discovery and inference, taking advantage of their ability to capture complex dependencies and relational structures in graph data \cite{zhao2024twist, chen2025revolutionizing, job2025hebcgnn}. Wang \emph{et al.}~\cite{wang2023incremental} designed a root cause analysis framework that performs causal learning using a Variational Graph Autoencoder (VGAE) to incrementally embed and update state-invariant and state-dependent causal relationships, enabling accurate and dynamic identification of root causes in evolving system states. GNNs’ causal learning ability was analyzed using a synthetic dataset with controlled causal relationships and Gao \emph{et al.}~\cite{gao2024rethinking} proposed a lightweight module to enhance their causal modeling, validated through experiments on synthetic and real-world data. Sui \emph{et al.}~\cite{sui2024enhancing} introduced \textit{CAL+}, a causal-based GNN framework using backdoor adjustment and attention mechanisms to separate causal and shortcut features, improving out-of-distribution generalization in graph classification. GNNs were also used to address hidden confounders and spillover effects in causal effect estimation from networked data, and Sui \emph{et al.}~\cite{sui2024invariant} introduced the Invariant Graph Learning (IGL) framework to learn environment-invariant representations, improving generalization across multiple environments.

\section{Preliminaries}
The foundations of our framework are causal learning and graph neural networks, which are briefly discussed in the following subsections.

\subsection{Causal Learning}

Causal learning is used to uncover cause-effect relationships from data, moving beyond correlations to identify how changes in one variable affect another. Structural Causal Models (SCMs) provide a formal framework to represent these relationships by modeling effects $E$ as deterministic functions of their direct causes $C$ and independent noise variables. This forms a causal graph $C \rightarrow E$, where edges represent causal influences, supporting intervention analysis \cite{peters2017elements}. 

Interventions in causal models involve modifying parts of the data-generating process to study their impact. A hard intervention sets a variable to a fixed value and a soft intervention alters the functional relationship or noise affecting the variable. These interventions produce new distributions that differ from observational data, enabling the identification of causal effects not captured by ordinary observations \cite{peters2017elements}. 

Counterfactual reasoning extends this further by considering alternate hypothetical outcomes for the same observed case. This is done by updating the noise variables based on observed outcomes, without changing the causal mechanisms. From this, SCMs enable us to ask `what if' questions about alternative actions, based on the individual’s observed outcome.

Fig.~\ref{fig:process} illustrates this process in four subfigures: (a) shows the basic observational SCM where \textit{C} influences effect \textit{E} through the function $f{_E}(C,N{_E})$, without any intervention. (b) shows a hard intervention where \textit{E} is set to a fixed value $e*$ and an explicit intervention node $do(E:=e*)$, indicating an external manipulation. (c) presents a soft intervention where the functional relationship is altered to $g{_E}(C) + \tilde{N}_E$, indicating a change in how \textit{E} depends on \textit{C} or its noise. (d) shows counterfactual inference, where based on an observed outcome ($B=1, T=1 \implies N_B=1$), the noise variable is updated without changing the causal structure. This supports reasoning about possible outcomes under alternative interventions, such as \textit{do(T=0)} \cite{peters2017elements}.

\subsection{Graph Neural Networks}

Graph Neural Networks (GNNs) are neural models designed for graph-structured data, where node representations are refined through a message-passing mechanism that aggregates information from neighbouring nodes. As a result, GNNs are able to learn embeddings that capture local graph structure and node attributes. These embeddings are then used for downstream tasks such as node classification (predicting node labels) and graph classification (predicting graph labels). Some of the common GNN architectures include Graph Convolutional Networks (GCN) \cite{kipf2016semi}, Graph Attention Networks (GAT) \cite{velivckovic2017graph} and Graph Sample and Aggregation (GraphSAGE) \cite{hamilton2017inductive}. 

For a graph $G=(V,E)$ with node features $X$, where $V$ is the set of nodes and $E$ is the set of edges, the representation of each node is updated through a sequence of \textit{AGGREGATE} and \textit{COMBINE} operations at each layer. Beginning with $H^0=X$, each layer updates node representations by aggregating neighbour information and combining it with the node's current state as shown in Eq.~\ref{eq-GNN} \cite{alma991006690901904691}.

\begin{equation} \label{eq-GNN}
\begin{aligned}
a_v^k &= \text{AGGREGATE}^k\left( \left\{ H_u^{k-1} : u \in \mathcal{N}(v) \right\} \right) \\
H_v^k &= \text{COMBINE}^k\left( H_v^{k-1}, a_v^k \right)
\end{aligned}
\end{equation}

\section{Problem Statement}
Given a graph $G = (V,E)$, where $V$ is the set of nodes and $E$ is the set of edges, with node feature matrix $ X \in \mathbb{R}^{|V| \times d} $ and label vector $ Y \in \mathbb{R}^{|V|}$, the objective is to learn a node classification function $f: (G, X, v) \rightarrow \hat{y}_v$ that predicts the label $\hat{y}_v$ of a node $v \in V$, based on its causally relevant neighbourhood.  To achieve this, the model applies targeted interventions on the neighbourhood structure of $v$ during training, enabling it to separate causal neighbours from spurious or confounding ones. Interventions on node connections and dependencies enable the model to learn invariant node representations that remain stable under structural perturbations and noise. Consequently, predictions are informed by stable causal mechanisms rather than spurious correlations, leading to improved reliability, robustness and generalization in node classification tasks.

\section{Methodology}
In this section, we describe the methodology of our proposed framework, detailing its core components. The proposed methodology is based on the principle of learning causally invariant node representations  through structured interventions on graph topology. Building on this principle, counterfactual neighbourhoods are generated to simulate alternative structural scenarios, enabling the disentanglement of causal relationships from spurious correlations. An attention-based edge scoring mechanism is used to quantify causal relevance and to guide adaptive edge perturbations, where learnable importance masks selectively adjust connections based on their estimated influence. This integrated design facilitates targeted structural interventions and invariant representation learning, thereby enhancing robustness and causal understanding in graph-based reasoning tasks.

\begin{figure*}[ht]
    \centering
    \subfloat{%
        \includegraphics[width=0.96\textwidth]{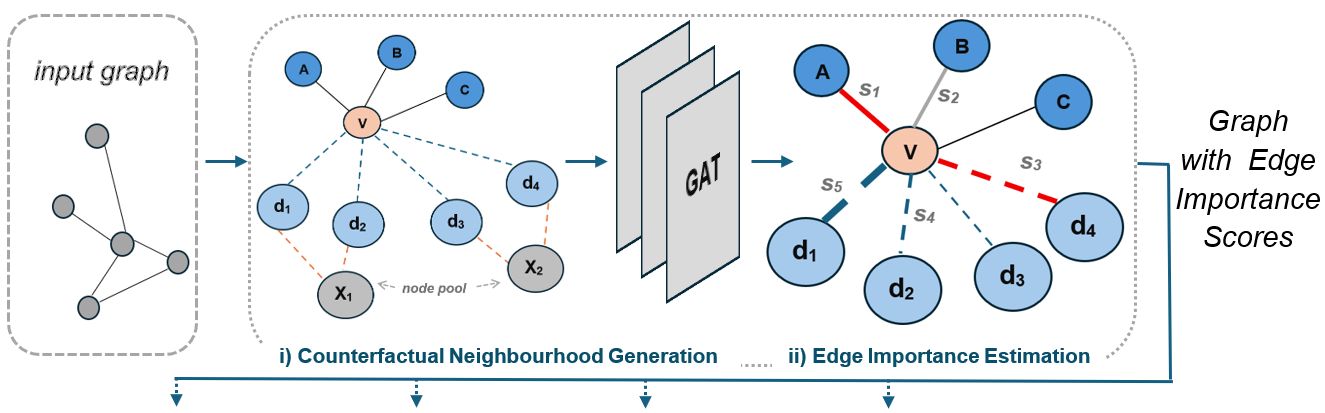}%
        \label{fig:design1}}
    \vspace{-1.1em}

    \subfloat{%
        \includegraphics[width=0.94\textwidth]{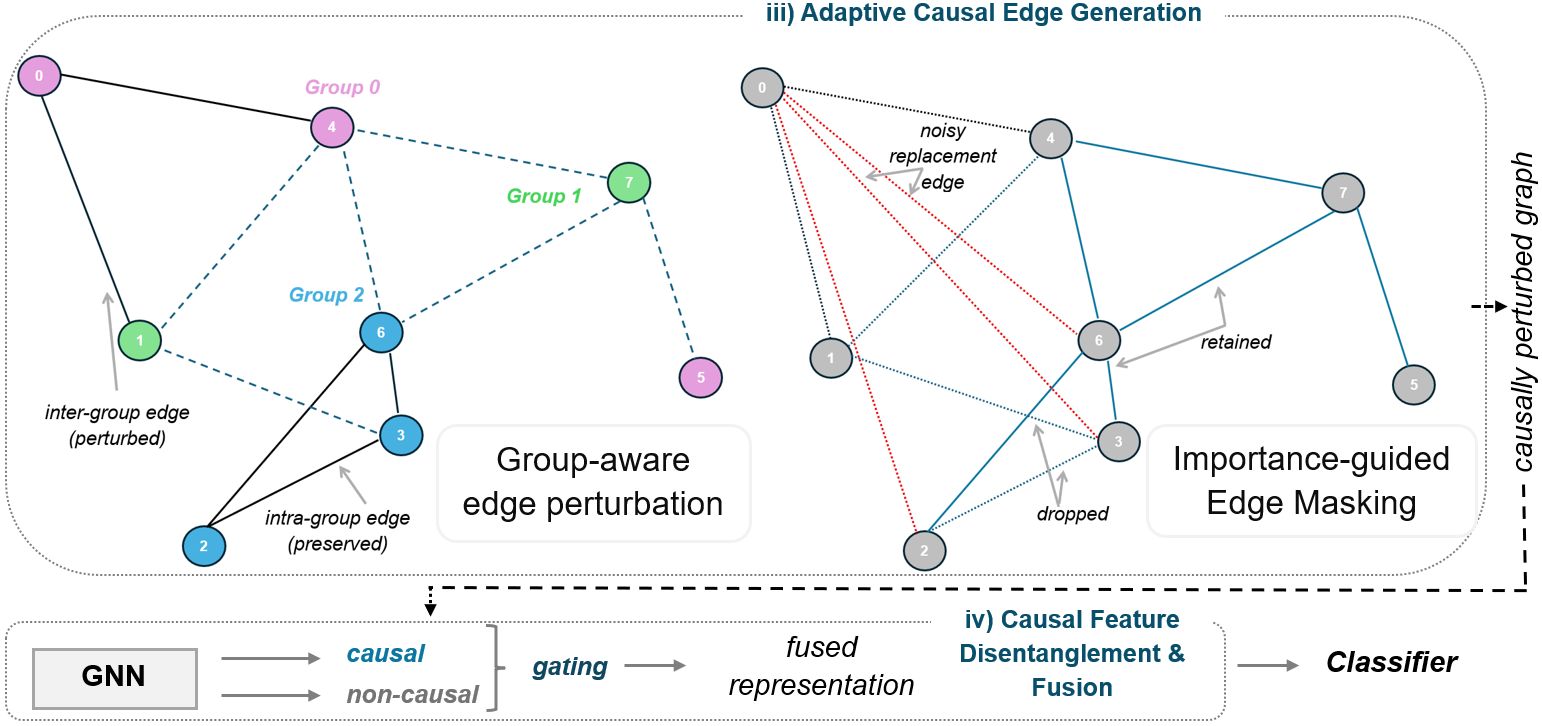}%
        \label{fig:design2}}

     \caption{CNL-GNN Architecture: (i) Enforces structure-invariant representations by perturbing neighbourhoods with dissimilar neighbours. (ii) Estimates edge relevance using attention to prioritize causal edges and guide structural perturbations. (iii) Selectively perturbs inter-group edges and masks low-importance edges to enhance causal robustness. (iv) Disentangles and fuses causal and non-causal features using gating for robust prediction.}
    \label{fig:Design}
\end{figure*}

\subsection{Causal Neighbourhood Learning with Graph Neural Networks (CNL-GNN)}

In this paper, we propose a causal graph learning framework designed to learn invariant node representation through counterfactual neighbourhood generation, adaptive edge perturbation and edge scoring. The framework is composed of four key components: (i) Counterfactual Neighbourhood Generation (ii) Edge Importance Estimator (iii) Adaptive Causal Edge Generator and (iv) Causal Feature Disentanglement and Fusion. Counterfactual Neighbourhood Generation, inspired by Pearl's causal models and counterfactual reasoning \cite{pearl1997causation, pearl2002reasoning}, simulates alternative neighbourhoods to approximate counterfactual outcomes. Earlier works have successfully used counterfactual reasoning for disentangling causal influences from confounders \cite{liu2024edvae, gao2025causal}. The Edge Importance Estimator identifies edges with causal influence on the target using attention mechanisms \cite{sui2024enhancing}, which effectively highlight causality and mitigate spurious relationships. In causal inference, a do-intervention refers to an operation that sets a variable to a fixed value, denoted as $do(X=x)$, removing the influence of its parents in the causal graph and simulating an intervention \cite{pearl2010introduction}. The Adaptive Causal Edge Generator, based on the do-intervention framework, performs graph perturbations according to the estimated causal effects, enabling the model to be more robust compared to random perturbations \cite{ahuja2023interventional, squires2023linear}. The final component disentangles node embeddings into contextual and object features, enabling invariant feature learning and robustness to confounding \cite{hu2025graph, zhu2025causal}. Each component addresses key challenges in causal graph learning such as confounding, spurious correlations and distribution shifts, resulting in a robust and generalizable classification framework. An overview of the architecture is illustrated in Fig.~\ref{fig:Design}. 

The methodology is detailed in Algorithm~\ref{alg:CNLGNNalg}. It begins with initializing all model components (line 1). For each node $v$ in graph $G$, a set of counterfactual neighbours $\mathcal{N}_c(v)$ is selected and used to construct a counterfactual graph $\tilde{G}_c$ (lines 3-6). Using $G$ and node features $X$, edge importance scores $S$ are computed to indicate edge relevance (line 7, Algorithm~\ref{alg:CNLGNNalgB}). Groups $\mathcal{G}$ are assigned by a detector to guide perturbations (line 8). A perturbed graph $\tilde{G}$ is generated by modifying $G$ based on $S$ and $\mathcal{G}$ (line 9). Node features $X$ are perturbed to obtain $\tilde{X}$ (line 10), introducing variation for robustness. The perturbed features and graph are input to a \textit{GATv2} model to generate node embeddings $x$ (line 11). A feature gate splits $x$ into context ($x_c$) and object ($x_o$) features (line 12), which are separately processed by two GNN branches (lines 13-15) and fused using a learned gating parameter $\alpha$ to obtain $x_f$ (line 16). Counterfactual features are generated by intervening on $x_c$ and $x_o$ enabling causal learning (line 17), followed by loss computation and parameter updates (line 18). Each of these components are described in detail in the following subsections.

\begin{algorithm}[H]
\caption{Causal Neighbourhood Learning with GNN (CNL-GNN)} \label{alg:CNLGNNalg}
\label{alg:1}
\begin{algorithmic}[1]
\Require Graph $G = (V, E)$, features $X$, epochs $T$
\Ensure Trained model

\State Initialize model parameters
\For{epoch $= 1$ to $T$}
    \ForAll{node $v \in V$}
        \State $\mathcal{N}_c(v) \gets$ sample $k$ dissimilar non-neighbours 
        \State Construct counterfactual graph $\tilde{G}_c$ 
    \EndFor
    \State $S \gets$ \textit{EstimateEdgeImportance}$(G, X)$
    \State $\mathcal{G} \gets$ group assignments from pretrained detector
    \State $\tilde{G} \gets \mathcal{M}_i(\mathcal{P}_g(G, \mathcal{G}), S)$ \Comment{Edge perturbation and masking}
    \State \textbf{Feature perturbation:} $\tilde{X} \gets \textit{PerturbFeatures}(X)$
    \State \textbf{Node encoding:} $x \gets \text{GATv2}(\tilde{X}, \tilde{G})$
    \State \textbf{Feature splitting:} $x_c \gets \text{FeatureGate}(x)$; \quad $x_o \gets x - x_c$
      \State \textbf{Branch processing:} 
        \State \quad $x_c \gets \text{ContextGNN}(x_c, \tilde{G})$
        \State \quad $x_o \gets \text{ObjectGNN}(x_o, \tilde{G})$
    \State \textbf{Feature fusion:} $x_f \gets \alpha \cdot x_c + (1 - \alpha) \cdot x_o$ 
   \State \textbf{Counterfactual intervention:} $(\tilde{x}_c, \tilde{x}_o) \gets \textit{GenerateCounterfactuals}(x_c, x_o)$

    \State Compute losses and update parameters
\EndFor \\
\Return model parameters
\end{algorithmic}
\end{algorithm}

\begin{algorithm}[H]
\caption{\textit{EstimateEdgeImportance}$(G, X)$} \label{alg:CNLGNNalgB}
\begin{algorithmic}[1]
\Require Graph $G$, features $X$
\Ensure Scores $S$
\State $h \gets \text{LinearProjection}(X)$ \Comment{project node features}
\ForAll{$(i, j) \in E$}
    \State $s_{ij} \gets \text{LeakyReLU} \left( (h_i \odot a) + (h_j \odot a) \right)$ \Comment{apply attention-weighted sum}
\EndFor
\State $S \gets \text{SoftmaxNormalize}(s, j)$ \Comment{j: target nodes}\\
\Return $S$
\end{algorithmic}
\end{algorithm}

\subsubsection{Counterfactual Neighbourhood Generation}

\textit{Counterfactual Neighbourhood Generator (CNG)} is designed to simulate causal interventions by perturbing the local graph structure around each node. For each target node, CNG generates perturbed neighbourhoods by sampling alternative neighbours using one of three strategies: (a) \textit{Random}: nodes sampled uniformly at random from the graph; (b) \textit{Similar}:  nodes with similar feature representations; and (c) \textit{Dissimilar}: nodes with dissimilar features. After empirical evaluation, the \textit{Dissimilar} strategy was selected as the default owing to its consistent performance across datasets. 

For each sampled set of counterfactual neighbours, a modified graph is constructed by bi-directionally adding edges between the target node and the sampled nodes, forming a counterfactual neighbourhood. This structural intervention simulates alternative causal scenarios, enabling the model to generalize under varying neighbourhood configurations as shown in Eq.~\ref{eq-CNG}. Here, $G$ is the original graph, $\tilde{G}$ is the counterfactual graph with perturbed neighbourhoods, $V$ is the set of nodes, and $\mathcal{N}_c(v)$ denotes the set of sampled counterfactual neighbours for node $v$. Edges between $v$ and $u$ are added bi-directionally to simulate structural interventions. Exposing the model to these synthetic structural variations enables CNG to enforce the learning of representations that are invariant to local structural changes. This approach encourages the model to focus on stable, causal patterns instead of overfitting to spurious correlations in the original neighbourhood.

\begin{equation} \label{eq-CNG}
\tilde{G} = G + \sum_{v \in V} \sum_{u \in \mathcal{N}_c(v)} \big( (v, u) + (u, v) \big)
\end{equation}

A contrastive loss is used to ensure consistency between the original and counterfactual graph representations by minimizing the difference in their predictions. This regularization enables the model to learn node features that are robust to confounding factors and capable of generalizing across different structural settings.

\subsubsection{Edge Importance Estimation}
The second component of the framework focuses on distinguishing causal edges from spurious or confounded ones through \textit{Edge Importance Module (EIM)}. This module estimates the relevance of each edge using an attention mechanism. For each edge $(i,j)$, an importance score $s_{ij}$ is computed using a learned neural attention function conditioned on the features of the connected nodes as shown in Eq.~\ref{eq-EIS}.

\begin{equation} \label{eq-EIS}
s_{ij} = \text{LeakyReLU} \left( \langle \mathbf{a}, W h_i \rangle + \langle \mathbf{a}, W h_j \rangle \right)
\end{equation}

These scores are used to guide structural perturbations and representation learning by assigning relative importance to edges. The model gives higher priority to causally informative connections and probabilistically perturbs or reduces the influence of less relevant or spurious edges. This selective weighting forms the basis for downstream structural modifications, ensuring that the model focuses on meaningful causal relationships. Edge relevance is quantified by EIM to guide the framework toward learning more robust and causally reliable graph structures. 

\subsubsection{Adaptive Causal Edge Generator}
In this stage, an Adaptive Causal Edge Generator is employed to selectively perturb the graph structure during training to regularize the model and prevent overfitting to spurious patterns. This process consists of two steps: (i) \textit{Group-aware Edge Perturbation:} A group detector is used to identify latent groups in the graph. Edges connecting nodes that belong to the same group are preserved, and edges between different groups are probabilistically perturbed, simulating local causal consistency. (ii) \textit{Importance-Guided Edge Masking:} Edge scores from the \textit{EIM} are used to selectively drop or retain edges based on their learned causal relevance. Edges with low importance scores are either removed or replaced with noisy alternatives.

This two-level strategy balances the preservation of meaningful structure by keeping important connections and removing noisy ones. Edges in the same group are assigned higher importance scores and are more likely to be retained, whereas edges between groups are treated as potentially spurious and perturbed with higher probability. Additionally, Gaussian noise is added to edge weights during training to further regularize the model and prevent overfitting. These operations as shown in Eq.~\ref{eq-EG} generate a perturbed graph $\tilde{G}$ that is used to train the model jointly with the original graph. This component improves robustness to spurious correlations and mitigates confounding variables, resulting in reliable causal node classification.

\begin{equation} \label{eq-EG}
\tilde{G} = \mathcal{M}_i\big(\mathcal{P}_g(G, \mathcal{G}), S\big)
\end{equation}

\noindent
where:
\begin{itemize}
    \item $G$ is the original graph,
    \item $\mathcal{G}$ represents the identified latent groups,
    \item $\mathcal{P}_g(G, \mathcal{G})$ applies group-aware edge perturbations on $G$ based on $\mathcal{G}$,
    \item $S$ denotes the edge importance scores from \textit{EIM},
     \item $\mathcal{M}_i(\cdot, S)$ applies edge masking based on $S$,
    \item $\tilde{G}$ is the perturbed graph used jointly with $G$ for training.
\end{itemize}

\subsubsection{Causal Feature Disentanglement and Fusion}
The final stage of the framework focuses on disentangling and integrating stable causal signals and unstable non-causal features for improved prediction. Two separate GNN modules generate disentangled latent representations, where $x{_c}$ captures stable, invariant (causal) features, and $x{_o}$ captures unstable, spurious features. Orthogonality loss is applied to reduce feature redundancy and mutual information losses minimize shared information, reinforcing their disentanglement. This separation is important because real-world node features often contain both causal and non-causal information, and treating them as a single embedding can negatively impact interpretability and performance. 

\begin{equation} \label{eq-FG}
x_{f} = \alpha \cdot x_c + (1 - \alpha) \cdot x_o
\end{equation}

The two representations are fused through a learnable gating mechanism that assigns node-specific weights, parameterized by a gating coefficient $\alpha$ computed from the concatenated representations $[x{_c}, x{_o}]$, as shown in Eq.~\ref{eq-FG}. The fused representation $x{_f}$ is used for final classification. This adaptive fusion allows the model to use both representations based on context and keeps a strong focus on causal features.
Adaptive causal edge perturbations from earlier stages further help in regularizing this process by exposing the model to perturbed graph structures during training. Overall, this module enhances prediction performance and generalizability under distribution shifts and structural noise.

\section{Experiments}

As discussed earlier, traditional GNNs are susceptible to spurious correlations, entangled causal and non-causal features and dependence on noisy or incomplete graph structures. To address these challenges, a causal graph learning framework based on Causal Neighbourhood Learning (CNL) is designed, which employs counterfactual reasoning and causal interventions to extract invariant and robust node representations. We evaluate the effectiveness of this approach on multiple datasets through experiments designed to test robustness, reduce spurious correlations and improve generalization under distribution shifts. Specifically, we aim to address the following research questions:

\textit{Research Questions}
\begin{enumerate}
    \item \textbf{RQ1.} What is the impact of counterfactual neighbourhood generation on node classification performance across different graph datasets?

    \item \textbf{RQ2.} What role do adaptive edge perturbations guided by learned causal relevance play in mitigating confounding variables during node classification?

    \item \textbf{RQ3.} How effectively does the proposed causal graph learning framework generalize under distributional shifts?
\end{enumerate}

\subsection{Experiment Design}
To address these research questions, we evaluate our model on multiple datasets with varying structural properties and noise levels. We compare our model against state-of-the-art GNN methods and conduct ablation studies to isolate the contribution of specific components in our framework, highlighting how each component supports building a more robust model. 

\subsection{Experimental Settings}
The experiments were conducted on Ubuntu 22.04 using Visual Studio Code 1.84, Python 3.11 and PyTorch, with an NVIDIA GeForce RTX 3070 Ti GPU. The model was trained for 20 epochs with a learning rate of \textit{1e-3}, with early stopping applied to avoid overfitting.

\subsection{Datasets}

\textbf{Citation:} \textit{Cora}, \textit{Citeseer} and \textit{PubMed} datasets are citation networks commonly used in graph classification research \cite{velivckovic2017graph, kipf2016semi, wu2024graph, xia2023learning, job2025causal}. In these datasets, nodes correspond to scientific publications and edges represent citation relationships between them. Each dataset categorizes the papers to 7, 6 and 3 classes respectively.  \\

\textbf{Twitch:} This social network dataset represents Twitch streamers as nodes, with edges representing social connections between them. Each node has a label denoting the type of content the streamer creates. The dataset consists of multiple regional subgraphs such as \textit{DE} (Germany), \textit{FR} (France), \textit{RU} (Russia), \textit{PT} (Portugal), \textit{EN} (English) and \textit{ES} (Spain) \cite{rozemberczki2021multi}. Twitch has been used in recent research on graph learning \cite{wu2024graph, hasan2025language} highlighting its relevance for tasks such as node classification.

\renewcommand{\arraystretch}{1.3}
 \begin{table}[h]

\caption{Overview of Datasets used in the Study}

 \centering
{\fontsize{10}{10}\selectfont
  \begin{tabular}
  {|p{1.8cm}|p{1.4cm}|p{1.4cm}|p{1.4cm}|}
  \hline
      \textbf{Dataset}   & \textbf{ \# Nodes } &  \textbf{\# Edges } & \textbf{\# Classes} \\
   
\hline
Cora  &  2708  & 10556  &   7 \\
Citeseer    & 3327  &  9104 & 6 \\
PubMed  &  19717 & 88648  & 3 \\
Twitch-DE    & 9498  &  315774 & 2 \\
Twitch-EN    & 7126  &  35324 & 2 \\
Twitch-PT    & 1912  &  31299 & 2 \\
Twitch-ES    & 4648  &  59382 & 2 \\
Twitch-RU    & 4385  &  37304 & 2 \\
Twitch-FR    & 6549  &  112666 & 2 \\
\hline 
\end{tabular}
}
\label{tab:tabldatasets}
\end{table}

\subsection{Baseline Models}
\begin{itemize}
    \item Information-based Causal Learning (ICL) \cite{zhao2024twist}: The ICL framework uses mutual information optimization and causal disentanglement to transform statistical correlations into invariant causal representations, enhancing GNN robustness and interpretability under distribution shifts.
   
    \item Adaptive Causal Module and Causality-Enhanced (ACE) \cite{chen2025revolutionizing}: This framework extracts causal features and controls confounding effects through adaptive causal modules and backdoor adjustment to enhance GNN generalization and interpretability without retraining on out-of-distribution data.  
    \item Causal Intervention for Network Data (CaNet) \cite{wu2024graph}: CaNet enhances GNN robustness to out-of-distribution node-level shifts by using a causal inference–based objective that combines an environment estimator with a mixture-of-expert predictor to mitigate latent confounding bias. 
    \item Causality and Independence Enhancement (CIE) \cite{chen2023causality}: CIE is a causal framework that enhances generalization under distribution shifts in graph neural networks by separating causal and spurious features through backdoor adjustment, addressing mixed data biases without requiring bias-specific model designs.
    \item Graph Contrastive Invariant Learning (GCIL) \cite{mo2024graph}: GCIL is a graph contrastive learning framework that uses spectral augmentation and causal objectives to learn invariant node representations by mitigating non-causal and confounding factors.
    \item Causality-driven Adaptive Sparsity and Hierarchical Memory framework (CAM) \cite{chen2025cam}: CAM is a causal framework that enhances generalization under distribution shifts in GNNs by dynamically selecting causally relevant expert groups and modeling inter-layer dependencies to eliminate spurious correlations without requiring environment labels.
\end{itemize}


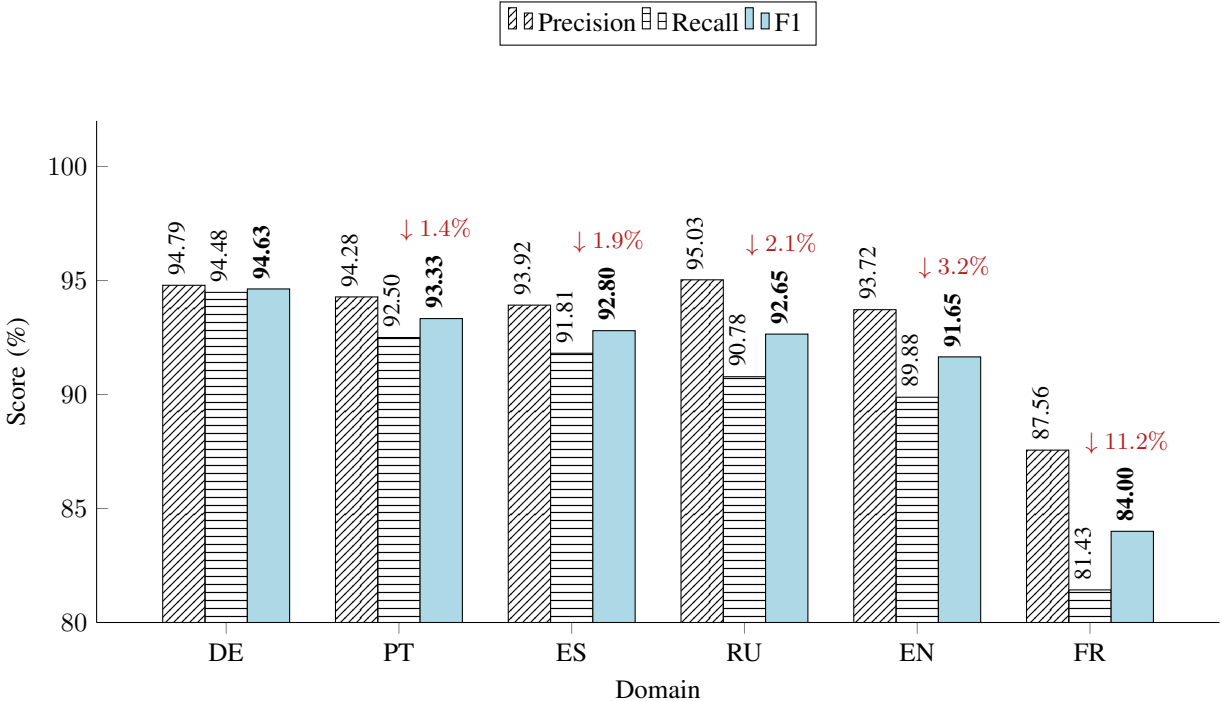
\begin{figure*}[htbp]
\centering
\begin{tikzpicture}
\definecolor{f1blue}{RGB}{173,216,230}      
\definecolor{precblue}{RGB}{200,220,240}    
\definecolor{recgray}{RGB}{220,220,220}     
\definecolor{dropred}{RGB}{180,30,30}       

\begin{axis}[
    ybar,
    width=\textwidth,
    height=0.5\textwidth,
    bar width=16pt,
    ymin=80, ymax=102,
    ylabel={Score (\%)},
    xlabel={Domain},
    symbolic x coords={DE, PT, ES, RU, EN, FR},
    xtick=data,
    enlarge x limits=0.15, 
    axis x line*=bottom,
    axis y line*=left,
    legend style={at={(0.5,1.15)}, anchor=south, legend columns=3},
]

\addplot+[ybar, draw=black, fill=precblue, pattern=north east lines, bar shift=-16pt] coordinates {  
    (DE,94.79) (PT,94.28) (ES,93.92) (RU,95.03) (EN,93.72) (FR,87.56)
};
\addplot+[ybar, draw=black, fill=f1blue!20, pattern=horizontal lines, bar shift=0pt] coordinates {  
    (DE,94.48) (PT,92.50) (ES,91.81) (RU,90.78) (EN,89.88) (FR,81.43)
};
\addplot+[ybar, draw=black, fill=f1blue, bar shift=16pt] coordinates {  
    (DE,94.63) (PT,93.33) (ES,92.80) (RU,92.65) (EN,91.65) (FR,84.00)
};

\node at (axis cs:DE,94.79)[xshift=-12pt, yshift=14pt, above, font=\small, rotate=90] {94.79};
\node at (axis cs:PT,94.28)[xshift=-12pt, yshift=14pt, above, font=\small, rotate=90] {94.28};
\node at (axis cs:ES,93.92)[xshift=-12pt, yshift=14pt, above, font=\small, rotate=90] {93.92};
\node at (axis cs:RU,95.03)[xshift=-12pt, yshift=14pt, above, font=\small, rotate=90] {95.03};
\node at (axis cs:EN,93.72)[xshift=-12pt, yshift=14pt, above, font=\small, rotate=90] {93.72};
\node at (axis cs:FR,87.56)[xshift=-12pt, yshift=14pt, above, font=\small, rotate=90] {87.56};

\node at (axis cs:DE,94.48)[xshift=4pt, yshift=14pt, above, font=\small, rotate=90] {94.48};
\node at (axis cs:PT,92.50)[xshift=4pt, yshift=14pt, above, font=\small, rotate=90] {92.50};
\node at (axis cs:ES,91.81)[xshift=4pt, yshift=14pt, above, font=\small, rotate=90] {91.81};
\node at (axis cs:RU,90.78)[xshift=4pt, yshift=14pt, above, font=\small, rotate=90] {90.78};
\node at (axis cs:EN,89.88)[xshift=4pt, yshift=14pt, above, font=\small, rotate=90] {89.88};
\node at (axis cs:FR,81.43)[xshift=4pt, yshift=14pt, above, font=\small, rotate=90] {81.43};

\node at (axis cs:DE,94.63)[xshift=20pt, yshift=14pt, above, font=\small, rotate=90] {\textbf{94.63}};
\node at (axis cs:PT,93.33)[xshift=20pt, yshift=14pt, above, font=\small, rotate=90] {\textbf{93.33}};
\node at (axis cs:ES,92.80)[xshift=20pt, yshift=14pt, above, font=\small, rotate=90] {\textbf{92.80}};
\node at (axis cs:RU,92.65)[xshift=20pt, yshift=14pt, above, font=\small, rotate=90] {\textbf{92.65}};
\node at (axis cs:EN,91.65)[xshift=20pt, yshift=14pt, above, font=\small, rotate=90] {\textbf{91.65}};
\node at (axis cs:FR,84.00)[xshift=20pt, yshift=14pt, above, font=\small, rotate=90] {\textbf{84.00}};

\node at (axis cs:PT,93.33)[xshift=14pt, above, dropred, font=\small, yshift=26pt] {$\downarrow 1.4\%$};
\node at (axis cs:ES,92.80)[xshift=14pt, above, dropred, font=\small, yshift=26pt] {$\downarrow 1.9\%$};
\node at (axis cs:RU,92.65)[xshift=14pt, above, dropred, font=\small, yshift=26pt] {$\downarrow 2.1\%$};
\node at (axis cs:EN,91.65)[xshift=14pt, above, dropred, font=\small, yshift=26pt] {$\downarrow 3.2\%$};
\node at (axis cs:FR,84.00)[xshift=14pt, above, dropred, font=\small, yshift=26pt] {$\downarrow 11.2\%$};

\legend{Precision, Recall, F1}
\end{axis}
\end{tikzpicture}
\caption{Precision, Recall and F1 evaluation under distribution shifts across Twitch domains. Arrows indicate relative F1 drop compared to the \textit{DE} domain.}
\label{fig:TwitchResults}
\end{figure*}

\subsection{Experiment Results}
This section provides an overview and analysis of the experimental findings. \\
\textit{Metrics.} The model is evaluated using 5-fold cross-validation, with the F-score as the primary metric. The F-score is particularly appropriate for node classification tasks because it balances precision and recall, addressing the challenges presented by imbalanced class distributions.

\renewcommand{\arraystretch}{1.3}
\begin{table}[h]

\fontsize{9}{9}\selectfont
\caption{F-Scores for node classification tasks \label{tab:tableResults}}
  \centering
  {\fontsize{9}{10}\selectfont
   \begin{tabular}  {p{2.3cm} p{0.9cm} p{0.9cm} p{0.9cm} p{0.9cm} }
  \hline

    \textbf{Model} &    Twitch  & Cora & Citeseer & PubMed  \\
    \hline
    GCN-ICL \cite{zhao2024twist} &  95.98 &  73.94 &  42.71 & 77.91  \\
    GAT-ICL \cite{zhao2024twist} &  96.71 &  74.47 &  51.76 & 84.15  \\
    ACE-GCN \cite{chen2025revolutionizing} & 64.50 & 61.90 &  35.93 & 73.03 \\ 
    ACE-GAT \cite{chen2025revolutionizing} &  68.12 &  72.37 &  58.88  & 78.67\\  

    CaNet \cite{wu2024graph} &  66.17 &  84.32 &   73.70 &  87.19 \\ 
    CIE \cite{chen2023causality} & 38.08 & 83.36  &  67.34  &  84.71 \\ 
    GCIL \cite{mo2024graph} &  61.97 &  71.59 &   42.64 &  76.66 \\ %
    CAM \cite{chen2025cam} & 65.99 & 90.87  &  79.75  &  88.08 \\ 
    \textbf{\textit{CNL-GNN}} (ours) &  \textbf{99.41} &  \textbf{93.51} &   \textbf{86.87} &  \textbf{90.23} \\

    \hline
    
\end{tabular}
}
\end{table}

Table~\ref{tab:tableResults} shows the F1-scores of \textit{CNL-GNN} compared to several state-of-the-art baselines on four datasets: \textit{Twitch}, \textit{Cora}, \textit{Citeseer} and \textit{PubMed}. \textit{CNL-GNN} consistently achieves the highest performance across all datasets, with F1-scores of 99.41 (Twitch), 93.51 (Cora), 86.87 (\textit{Citeseer}) and 90.23 (\textit{PubMed}). Other models such as \textit{ICL}, \textit{ACE}, \textit{CaNet}, \textit{CIE}, \textit{GCIL} and \textit{CAM} show competitive but consistently lower scores compared to \textit{CNL-GNN}.

\subsection{Analysis of Results}

The results demonstrate that \textit{CNL-GNN} significantly outperforms existing state-of-the-art models across all datasets. For instance, on the \textit{Twitch} dataset, \textit{CNL-GNN} achieves an F1-score of 99.41, which is 2.7 \% higher than the next best model, \textit{GAT-ICL}. On \textit{Cora}, \textit{CNL-GNN} improves upon \textit{CAM} and \textit{CaNet} by approximately 2.6 and 9.2\% respectively. Similarly, on \textit{Citeseer}, the model outperforms \textit{CAM} by around 7\% and \textit{CaNet} by 13\%. These improvements confirm the robustness and effectiveness of the proposed model across diverse datasets and graph structures. 

The consistently high F1-scores achieved by \textit{CNL-GNN} across all datasets can be attributed to the effective combination of its core components, each designed to address specific challenges in causal and robust graph learning. The model's superior performance - such as 99.41 on \textit{Twitch}, 93.51 on \textit{Cora}, 86.87 on \textit{Citeseer}, and 90.23 on \textit{PubMed} - demonstrates the effectiveness of its design in learning stable and generalizable node representations.

These gains are the result of the model’s ability to simulate counterfactual neighbourhoods that expose it to alternative graph structures, thereby enforcing robustness to structural perturbations and reducing dependence on spurious patterns. This is particularly evident in datasets like \textit{Cora} and \textit{PubMed}, where small structural changes can otherwise lead to reduced performance. Additionally, by estimating edge importance and using these scores to control selective perturbation and masking, the model preserves causal edges and suppresses noisy or irrelevant connections. This selective regularization contributes significantly to the strong performance observed on \textit{Citeseer}, which is more sensitive to structural noise. Furthermore, disentangling causal from spurious features and adaptively fusing them enables the model to retain essential information while discarding noise, leading to consistently high predictive accuracy even under distributional shifts. 

Domain shift experiments were conducted to evaluate the robustness and generalization capability of \textit{CNL-GNN} across different \textit{Twitch} language domains. For these experiments, training was limited to two epochs to balance computational efficiency with sufficient learning for assessing cross-domain generalization. Fig.~\ref{fig:TwitchResults} provides a cross-domain evaluation of \textit{CNL-GNN} to assess its generalizability across multiple Twitch language domains. The model trained on the German (\textit{Twitch-DE}) subset is evaluated on five unseen domains (\textit{PT}, \textit{ES}, \textit{RU}, \textit{EN}, \textit{FR}). Performance remains consistently strong across all domains, though a gradual decline in F1-score is observed relative to the training domain (94.63\% on Twitch-DE). F1-scores range from 93.33\% (\textit{Twitch-PT}) to 84.00\% (\textit{Twitch-FR}), indicating varying levels of robustness under distribution shifts. These differences are accompanied by variations in precision and recall, providing insight into how the model adapts to linguistic and structural variations. Notably, the smallest F1 degradation (1.4\%) is observed for \textit{Twitch-PT}, while the largest drop (11.2\%) occurs for \textit{Twitch-FR}, highlighting the challenge of domain transfer across languages. This analysis validates the effectiveness of \textit{CNL-GNN} in maintaining high performance under distributional changes and demonstrates its generalization capability across diverse target domains.

\begin{figure*}[htbp]
\centering
\begin{tikzpicture}
\begin{axis}[
    width=\textwidth,
    height=0.5\textwidth,
    ybar,
    bar width=16pt,                  
    ymin=80, ymax=104,               
    ylabel={F1 Score (\%)},
    xlabel={Dataset},
    symbolic x coords={Cora, Citeseer, PubMed, Twitch},
    xtick=data,
    enlarge x limits=0.15,          
    legend style={at={(0.5,1.15)}, anchor=south, legend columns=5},
    nodes near coords,
    every node near coord/.append style={
        font=\small, black, rotate=90, anchor=west, yshift=2pt
    },
    cycle list={
        {fill={rgb:red,0.2;green,0.2;blue,0.4},  draw=black},  
        {fill=white, pattern=north east lines, draw=black},
        {fill=white, pattern=north west lines, draw=black},
        {fill=white, pattern=crosshatch, draw=black},
        {fill=white, pattern=dots, draw=black}
    },
]

\addplot+[bar shift=-32pt]  coordinates {(Cora,93.51) (Citeseer,86.87) (PubMed,90.23) (Twitch,99.41)};
\addplot+[bar shift=-16pt]  coordinates {(Cora,87.27) (Citeseer,81.33) (PubMed,85.11) (Twitch,98.49)};
\addplot+[bar shift=0pt]    coordinates {(Cora,86.93) (Citeseer,80.94) (PubMed,85.27) (Twitch,98.49)};
\addplot+[bar shift=16pt]   coordinates {(Cora,87.36) (Citeseer,81.80) (PubMed,85.92) (Twitch,98.39)};
\addplot+[bar shift=32pt]   coordinates {(Cora,87.67) (Citeseer,81.77) (PubMed,85.14) (Twitch,98.38)};

\legend{CNL-GNN, w/o CNG, w/o EIM, w/o group, w/o EIM+group}
\end{axis}
\end{tikzpicture}
\caption{F1 Scores across datasets (CNL-GNN vs. Ablated Variants). \textit{CNL-GNN} bars are shown in blue, ablations are white with hatch patterns.}
\label{fig:Ablation}
\end{figure*}
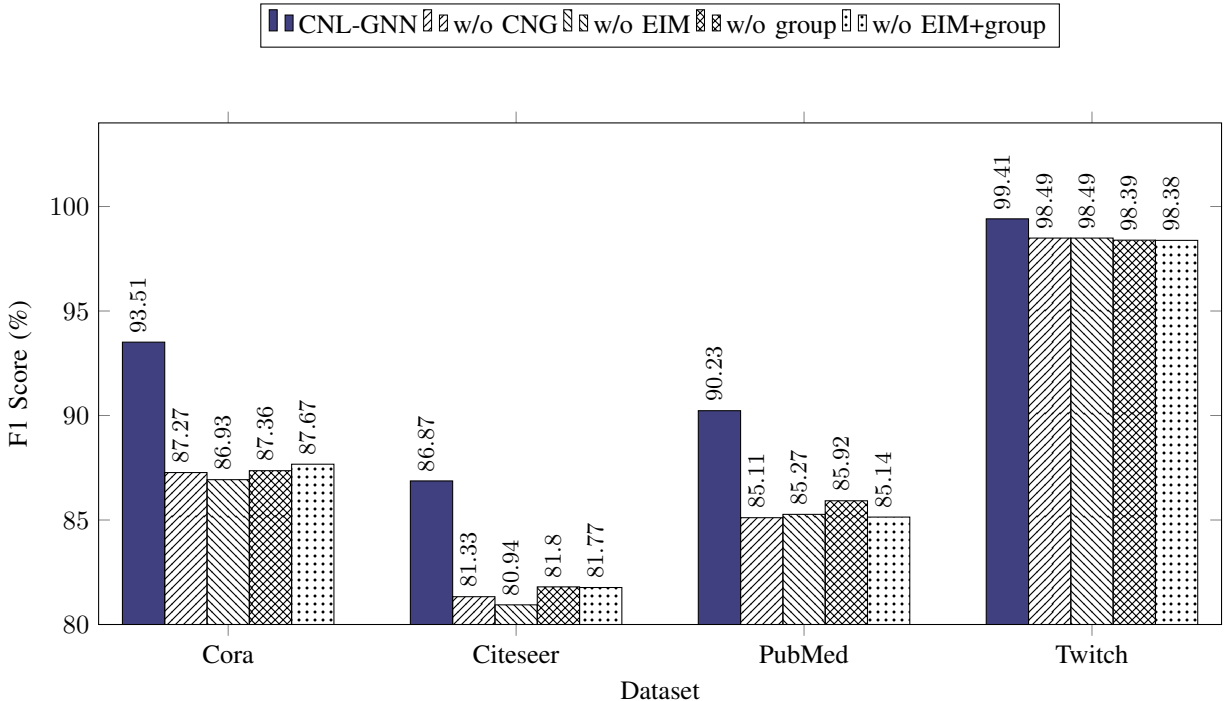

\subsection{Ablation Study}

An ablation study was conducted to evaluate the contribution of each component in the proposed model. The ablated variants are (i) \textit{w/o CNG}: exclusion of CNG module, (ii) \textit{w/o EIM }: exclusion of EIM, (iii) \textit{w/o group}: exclusion of Group-aware Edge Perturbation, (iv) \textit{w/o EIM+group}: exclusion of EIM and the group module. As shown in Fig.~\ref{fig:Ablation}, the original \textit{CNL-GNN} consistently outperforms all ablated versions across the four datasets - \textit{Cora}, \textit{Citeseer}, \textit{PubMed} and \textit{Twitch}.

The removal of any individual module leads to a noticeable decline in F1-scores, confirming their effectiveness. The most significant drop is observed when both EIM and the group modules are removed, highlighting their combined contribution. These results indicate that all components contribute positively to model performance and their integration yields the best results.

\subsection{Sensitivity Analysis}

\begin{figure*}[t]
\centering
\includegraphics[width=\textwidth]{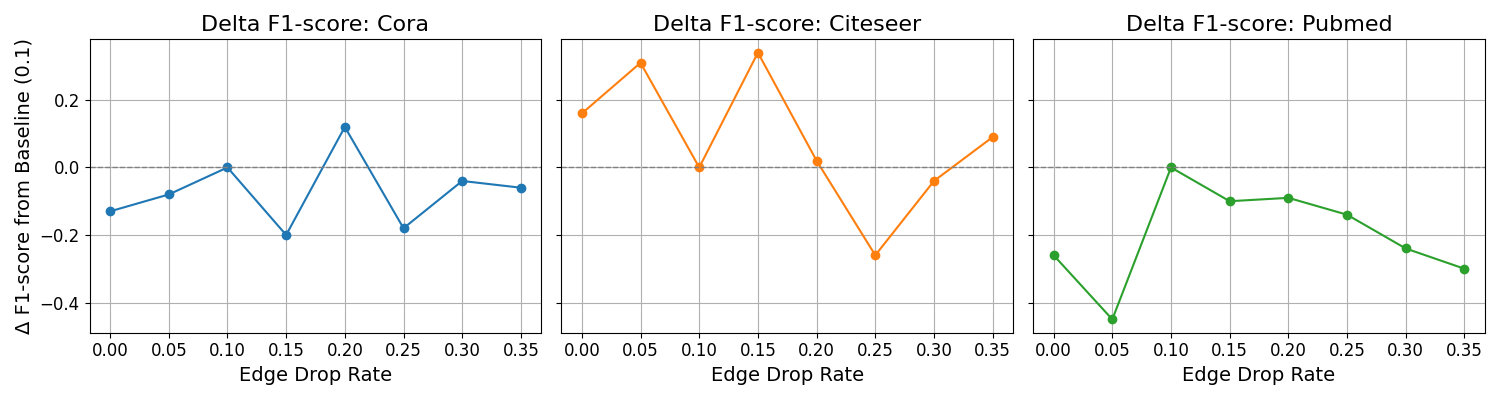}
\caption{Sensitivity Analysis: $\Delta$ F1-score vs. Edge Drop Rate for \textit{Cora}, \textit{Citeseer} and \textit{Pubmed}. Values indicate deviation from baseline at drop rate = 0.1.}
\label{fig:Sensitivity}
\end{figure*}

A sensitivity analysis was conducted for the model, with results shown in Fig.~\ref{fig:Sensitivity}. The plot represents the change ($\Delta$) in F1-score relative to the baseline value at an edge drop rate of 0.1 for three datasets. Positive values indicate improvements over the baseline, and negative values indicate a decline. This illustrates how the model responds to varying levels of structural perturbations in the graph. 

\textit{Cora} shows stable performance across all drop rates, with minimal F1-score fluctuation, indicating the model's robustness. \textit{Citeseer} exhibits slightly more variation, with a dip at 0.25, suggesting mild sensitivity. \textit{Pubmed} achieves its highest F1-score at 0.1 with marginal changes at other rates, indicating consistent performance. Overall, all three datasets demonstrate relatively low sensitivity to edge drop rate variations, with optimal results in the 0.1-0.2 range, supporting the choice of 0.1 as the default edge drop rate in our model.

\subsection{Summary}

Based on these findings, we provide the following answers to the research questions:

\textbf{RQ1.} \textit{CNL-GNN’s} Counterfactual Neighbourhood Generator (CNG) plays a crucial role in enhancing node classification performance by simulating structural interventions that generate perturbed neighbourhoods with dissimilar nodes. This mechanism forces the model to learn node representations invariant to local graph perturbations, effectively reducing overfitting to spurious correlations in the original neighbourhood structure. The consistent improvement in F1-scores across datasets such as \textit{Cora}, \textit{PubMed} and \textit{Twitch} validates the efficacy of CNG in capturing stable causal relationships in graph data.

\textbf{RQ2.} The adaptive edge perturbation strategy in \textit{CNL-GNN}, which employs importance scores from the Edge Importance Module (EIM), actively mitigates the influence of confounding variables by selectively preserving edges with high causal relevance and perturbing or masking less important ones. By combining group-aware edge preservation with importance-guided masking, this approach introduces controlled structural noise during training, preventing the model from overly depending on spurious edges. The ablation results and F1-scores across datasets demonstrate that this targeted approach effectively highlights key causal edges and improves model robustness.

\textbf{RQ3.} \textit{CNL-GNN} demonstrates effective generalization under distributional shifts, as demonstrated by its cross-domain evaluation on multiple \textit{Twitch} language domains. Trained solely on the German domain (\textit{Twitch-DE}), the model maintains strong performance when tested on five unseen domains (\textit{PT}, \textit{ES}, \textit{RU}, \textit{EN}, \textit{FR}). Although a gradual decrease in F1-score is observed relative to the training domain (94.63\%), the model achieves robust F1-scores ranging from 93.33\% (\textit{Twitch-PT}) to 84.00\% (\textit{Twitch-FR}). The varying degrees of performance drop (from 1.4\% up to 11.2\%) reflect the impact of domain shifts, including linguistic and structural differences. These results validate \textit{CNL-GNN}’s ability to maintain high predictive accuracy even when applied to unseen target domains, highlighting its robustness and adaptability to distributional changes in real-world settings.

\section{Conclusion}
In this work, we introduced Causal Neighbourhood Learning with Graph Neural Networks (CNL-GNN), a novel framework designed to improve node classification by incorporating causal interventions on graph structure. Our method generates counterfactual neighbourhoods and employs adaptive edge perturbation guided by learnable importance masking to separate causally relevant connections from spurious correlations. Experimental results across multiple datasets demonstrate that \textit{CNL-GNN} consistently achieves excellent performance compared to state-of-the-art GNN models by learning invariant node representations robust to local graph perturbations and distributional shifts. The Counterfactual Neighbourhood Generator enables the model to simulate structural variations that support invariance to noisy connections, thereby reducing overfitting. The adaptive edge perturbation strategy enhances robustness by selectively preserving key causal edges and masking confounding ones. Moreover, \textit{CNL-GNN} generalizes well across different domains, maintaining strong performance under distribution shifts, as validated in cross-domain evaluations on the \textit{Twitch} dataset.

Overall, these findings confirm that structural-level causal interventions combined with feature disentanglement are effective for learning robust and generalizable graph representations. This work extends causal graph learning beyond traditional feature-based methods, providing a pathway towards building robust GNN models in real-world applications. In the future, this method could be applied to dynamic graphs and more complex graph types to further improve its ability to generalize.


\bibliographystyle{IEEEtran}
\bibliography{causalgnnbib}

\end{document}